\documentclass{article}

\usepackage[final, nonatbib]{icbinb2021}

\usepackage[utf8]{inputenc} 
\usepackage[T1]{fontenc}    
\usepackage{hyperref}       
\usepackage{url}            
\usepackage{booktabs}       
\usepackage{amsfonts}       
\usepackage{nicefrac}       
\usepackage{microtype}      
\usepackage{xcolor}         

\usepackage{subcaption}
\usepackage{amsmath}
\usepackage{amssymb}
\usepackage{gensymb}
\usepackage{graphicx}
\usepackage{placeins}

\renewcommand{\b}{\boldsymbol}
\DeclareMathOperator*{\argmax}{arg\,max}

\title{An Underexplored Dilemma between Confidence and Calibration in Quantized Neural Networks}

\author{%
  Guoxuan Xia
  \\
  Imperial College London\\
  London, UK\\
  \texttt{g.xia21@imperial.ac.uk} \\
  \And
   Sangwon Ha \hspace{0.15cm}
   Tiago Azevedo \hspace{0.15cm}
   Partha Maji \\
   Arm ML Research Lab \\
   Cambridge, UK \\
\texttt{first.last@arm.com}
}

\begin{document}

\maketitle
\setcounter{footnote}{0}
\begin{abstract}
Modern convolutional neural networks (CNNs) are known to be overconfident in terms of their calibration on unseen input data. That is to say, they are more confident than they are accurate. This is undesirable if the probabilities predicted are to be used for downstream decision making. When considering accuracy, CNNs are also surprisingly robust to compression techniques, such as quantization, which aim to reduce computational and memory costs. 
We show that this robustness can be partially explained by the calibration behavior of modern CNNs, and may be improved with overconfidence. 
This is due to an intuitive result: low confidence predictions are more likely to change post-quantization, whilst being less accurate. High confidence predictions will be more accurate, but more difficult to change. Thus, a minimal drop in post-quantization accuracy is incurred. This presents a potential conflict in neural network design: worse calibration from overconfidence may lead to better robustness to quantization. We perform experiments applying post-training quantization to a variety of CNNs, on the CIFAR-100 and ImageNet datasets, and make our code publicly available.\footnote{\href{https://github.com/Guoxoug/PTQ-acc-cal}{\texttt{https://github.com/Guoxoug/PTQ-acc-cal}}}
\end{abstract}

\section{Introduction}\label{introduction}
We want predictive models to produce reliable uncertainty estimates, such that better informed decisions can be made based on their predictions. This is especially important in safety-critical applications such as autonomous driving \cite{kendall2017uncertainties} and medical diagnosis \cite{Kompa2021SecondON}. One common way of measuring the quality of predictive uncertainty is model calibration, which in the case of a classification task can be understood as how well the probabilities predicted by a model match its accuracy. For example, a well-calibrated model should be correct 70\% of the time, when it predicts a probability of 0.7 for the chosen class. Thus, for a $K$-class classification problem, we can define a \textit{perfectly calibrated} model \cite{DBLP:journals/corr/abs-2101-05397, pmlr-v70-guo17a} as,
\begin{equation}\label{cal}
    P(y=\hat y|P(\hat y|\b x; \b \theta) = p) = p, \quad p\in [0,1],
\end{equation}
where $\b \theta$ are the model parameters, $\b x$ is an input and $y \in \{\omega_k\}_{k=1}^K$ is the corresponding class label. $\hat y$ is the class predicted by that model and is given by $ \hat y = \argmax_\omega P(\omega|\b x; \b\theta)$, and $P(\hat y|\b x; \b \theta)$ is its corresponding predicted probability, or the model confidence for input $\b x$. In the case of a CNN, probabilities will come from the softmax after the final fully connected layer (or logits).
\\~\\
It has been previously shown that modern convolutional neural networks (CNNs), such as ResNets \cite{He2016DeepRL}, exhibit poor calibration, and in fact tend to be overconfident \cite{DBLP:journals/corr/abs-2101-05397, pmlr-v70-guo17a}. The model confidences are greater than the actual test accuracies achieved, making them unreliable. 
\\~\\
On the other hand, neural network quantization is an optimization technique where weights and activations are stored in a lower bit numerical format compared to what they were trained in, e.g. int8 vs fp32 \cite{jacob2017quantization, Krishnamoorthi, nagel2021white}. This can allow significant reductions in computational and memory costs, resulting in lower latency and energy consumption. Quantization is thus often important for deploying CNNs on resource limited platforms such as mobile phones. In this paper we examine post-training quantization (PTQ), where a fully-trained full-precision network is mapped to lower precision without further training. It has been noted that CNNs are robust to the noise introduced by quantization; however, to the best of our knowledge, research tends to focus on improving robustness \cite{nagel2021white, Alizadeh2020GradientR, Gholami2021ASO} rather than explaining why architectures seem to be inherently robust. There has been other research into understanding how compression methods affect accuracy \cite{hooker2021compressed}, however, it primarily focuses on the effect of pruning and does not consider predictive uncertainty. 
\\~\\
In this paper we highlight a novel insight, that links the above two concepts (calibration, quantization) in deep learning together. We find that
\begin{itemize}
    \item confidence and calibration are closely linked to the robustness of model accuracy to post-training quantization and,
    \item overconfidence can potentially improve the robustness of accuracy.
\end{itemize}

\section{How does calibration affect accuracy post quantization?}\label{idea}
We can consider the effect of post-training quantization on a CNN's activations and ultimately outputs as adding noise to the original floating point values \cite{Alizadeh2020GradientR, Yun2021DoAM}. Thus, we can separate two questions that determine the change in accuracy after quantization:
\begin{enumerate}
    \item How easy is it to change the predicted class?
    \item Given the prediction has changed, how will the model accuracy be influenced?
\end{enumerate}
In order for a prediction to change, for standard classification CNNs, the quantization noise needs to be sufficient to change the top logit. Intuitively, this suggests that less confident predictions will be easier to change, as their top logit will be closer to the other logits. We would also expect lower precision, and thus greater quantization noise, to result in more swapped top logits as well. We do not investigate the above in detail, as this would require accurately modelling the distribution of logits post-quantization for different architectures conditional on the input. However, we simply state that our empirical results support the intuition presented above.
\\~\\
The second question can be directly linked to the calibration of the model, as calibration tells us about the accuracy of predictions at a given confidence. For example, considering a well-calibrated model as defined in Equation \ref{cal}, for a prediction with confidence $P(\hat y|\b x; \b \theta) < 0.5$ the probability that it will be correct is also $ < 0.5$. This means that \textit{it is more likely to already be wrong}. If the prediction is in fact incorrect then it will not cause a decrease in the overall accuracy if it changes, since it will either change to another incorrect class, or to the correct class. 
\\~\\
It can now be seen how overconfidence, where the model is more confident than it is accurate, $ P(y=\hat y|P(\hat y|\b x; \b \theta) = p) < p$, may improve robustness to post-training quantization. For predictions with $P(y=\hat y)  > 0.5$, these are more likely to be correct in the first place, and so it would be better for these to stay the same post quantization. Thus, having a higher confidence would be beneficial. Conversely, for more easily swapped predictions with lower confidence, if the original accuracy of these is lower, then the change in accuracy post quantization will be less. 
\section{Experiments}
We present experimental results for ResNet56 and ResNet50 \cite{He2016DeepRL} trained on CIFAR-100 \cite{CIFAR} and ImageNet \cite{deng2009imagenet} respectively. Additional results for ResNet20 on CIFAR-100 and MobileNetV2 \cite{Sandler2018MobileNetV2IR} on ImageNet are available in Appendix \ref{appendix}. CIFAR models were trained using the regime specified by \cite{He2016DeepRL}, whilst ImageNet models use pretrained weights available from PyTorch\footnote{\href{https://pytorch.org/vision/stable/models.html}{\texttt{https://pytorch.org/vision/stable/models.html}}} \cite{NEURIPS2019_9015}.
\\~\\
For weights we use uniform per-channel symmetric quantization, where the quantization parameters are determined using minimum and maximum values. For activations we use uniform per-tensor asymmetric quantization, where the quantization parameters are found using PyTorch's default histogram based method that iteratively aims to minimize the mean squared error from quantization \cite{nagel2021white, NEURIPS2019_9015}. Batchnorm layers are folded into the preceding convolutional layer. This is a relatively standard scheme, as our aim is not to achieve the best performance, but to examine behavior as quantization noise varies.  Quantized inference is simulated in PyTorch using the existing backend for this purpose.\footnote{\href{https://github.com/pytorch/pytorch/blob/master/torch/ao/quantization/fake_quantize.py}{\texttt{https://github.com/pytorch/pytorch/blob/master/torch/ao/quantization/fake\_quantize.py}}}
\subsection{Model calibration before quantization}
\begin{figure}
    \centering
    \begin{subfigure}{0.49\linewidth}
        \centering
        \includegraphics[width=\linewidth]{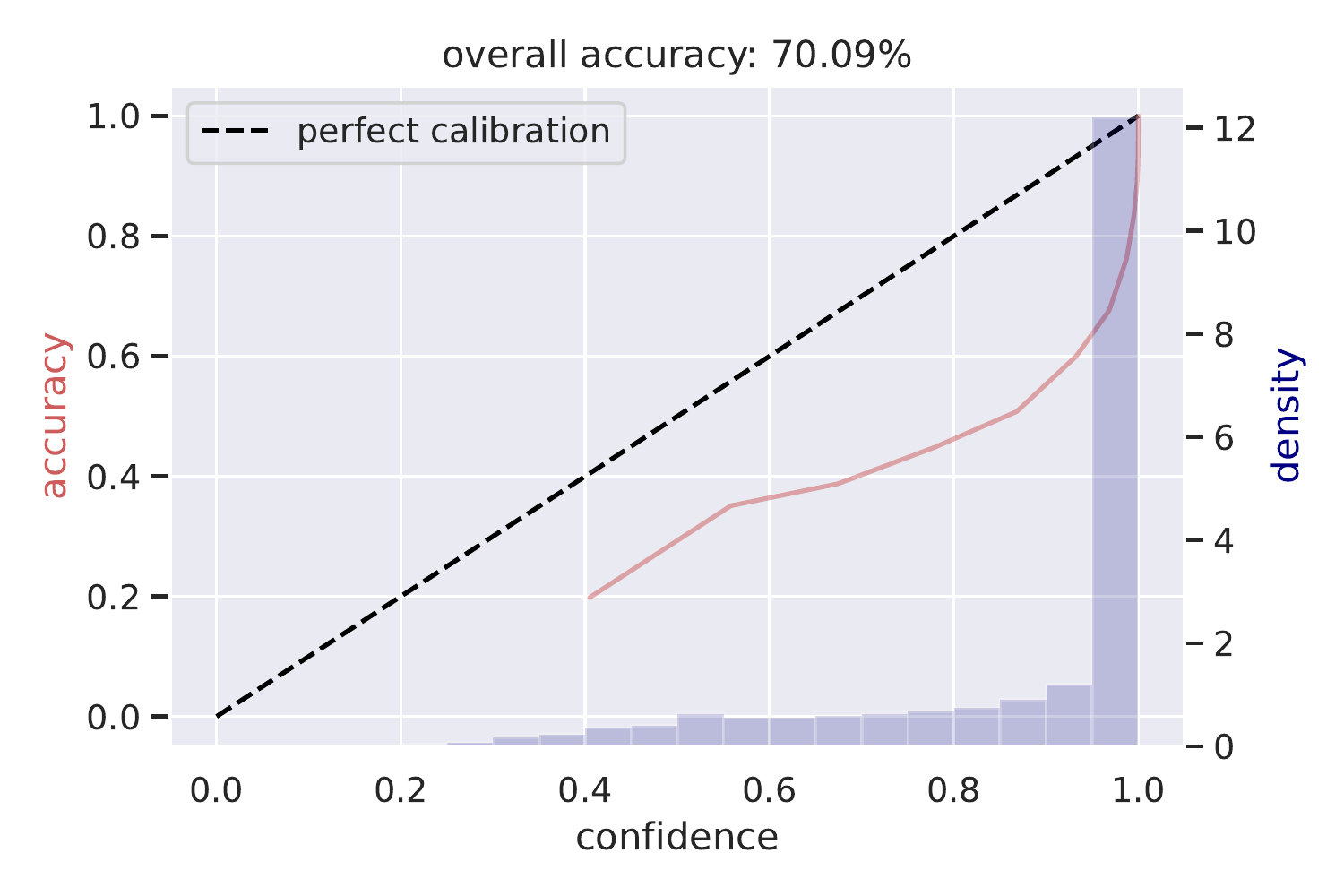}
        \caption{ResNet56, CIFAR-100}
    \end{subfigure}
    \begin{subfigure}{0.49\linewidth}
        \centering
        \includegraphics[width=\linewidth]{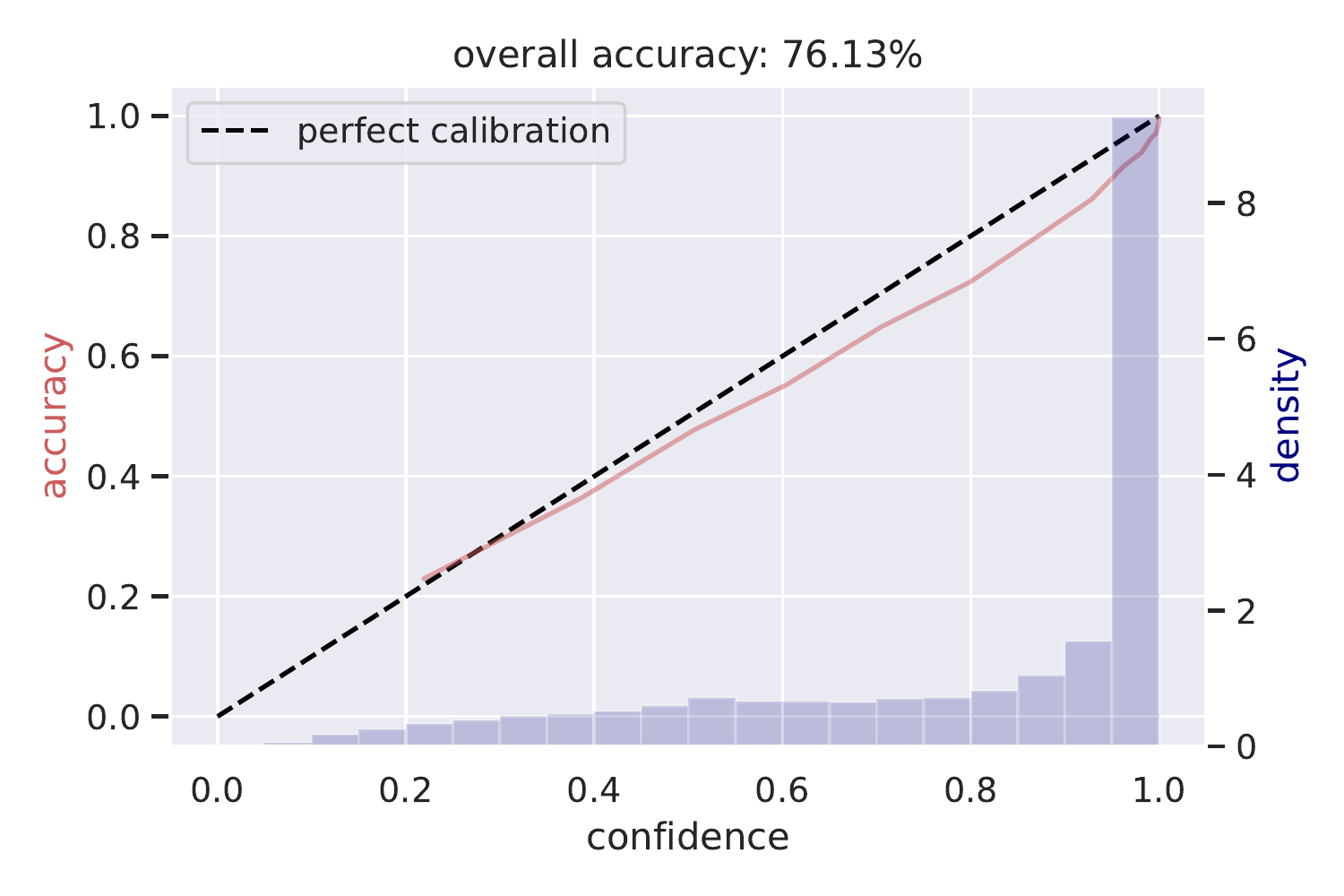}
        \caption{ResNet50, ImageNet}
    \end{subfigure}
    \caption{Reliability curves and confidence histograms for full precision (pre-quantization) networks on the corresponding test datasets. ResNet56 on CIFAR-100 is both very overconfident in terms of calibration (confidence $>$ accuracy), and highly confident in general. ResNet50 on ImageNet shows similar behavior but to a lesser extent.}
    \label{fig:1}
\end{figure}

Reliability curves plot accuracy against binned confidence \cite{pmlr-v70-guo17a, NiculescuMizil2005PredictingGP}, and so not only give an idea of the calibration error, but also of whether a model is over or under confident. Figure \ref{fig:1} shows reliability curves for floating-point models (pre-quantization), alongside histograms showing the distribution of confidence over the test datasets. It can be seen that ResNet56 on CIFAR-100 is very overconfident on the test data, with accuracy much lower than confidence. ResNet50 on ImageNet is better calibrated, but still overconfident as well. ResNet56 is also more confident overall compared to ResNet50, although both models have a large proportion of their predictions with confidence near $1.0$.
\subsection{Model accuracy after quantization}
\begin{figure}
    \centering
    \begin{subfigure}{0.49\linewidth}
        \centering
        \includegraphics[width=\linewidth]{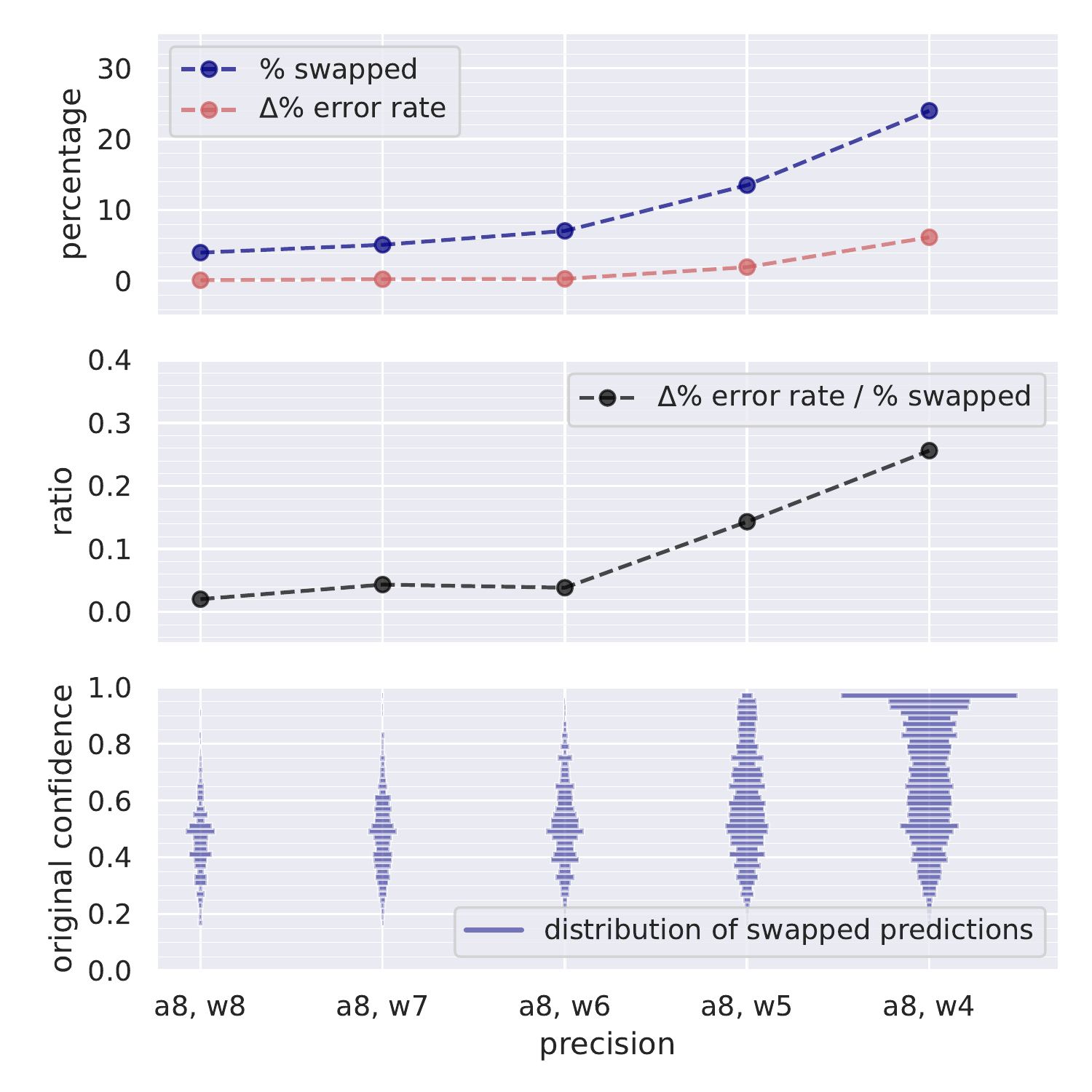}
        \caption{ResNet56, CIFAR-100}
    \end{subfigure}
    \begin{subfigure}{0.49\linewidth}
        \centering
        \includegraphics[width=\linewidth]{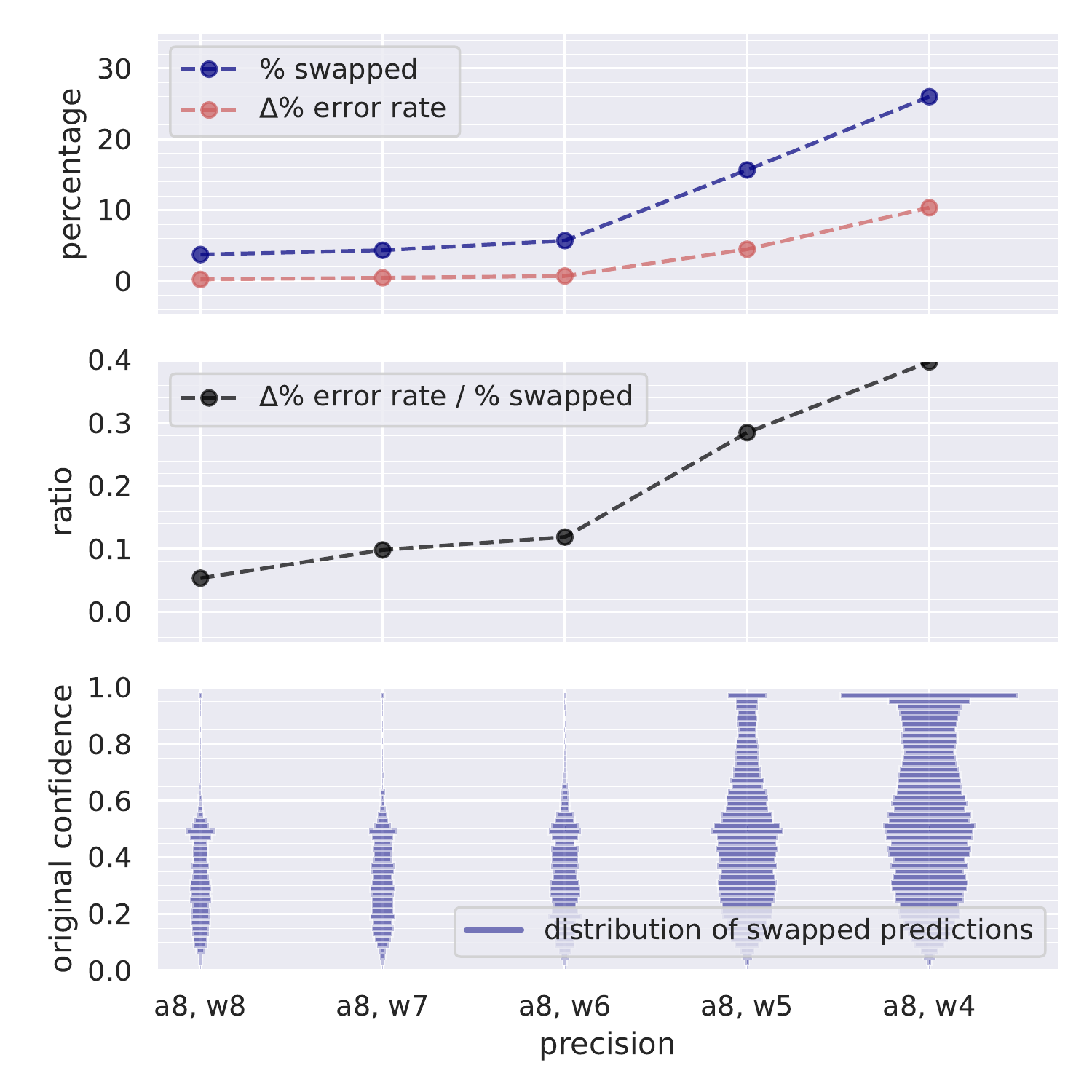}
        \caption{ResNet50, ImageNet}
    \end{subfigure}
    \caption{Going from floating point to quantized: percentage of swapped predictions, percentage change in error rate, their ratio, and (unnormalized) histograms of swapped predictions over confidence. These are plotted as weight (w) precision is decreased whilst activation (a) precision is held constant. Initially, as quantization noise grows the ratio stays low, as mainly low confidence (and low accuracy) predictions are swapped. The ratio starts to grow more rapidly, alongside the error rate, when higher confidence (and higher accuracy) predictions start to be swapped at higher noise levels.}
    \label{fig:2}
\end{figure}
Given knowledge of the calibration behavior of the networks (Figure \ref{fig:1}), we can explain the behavior of accuracy as quantization noise increases. Figure \ref{fig:2} shows, going from the floating point model to a quantized one, the percentage of swapped/changed predictions, the change in error rate ($1-$accuracy), the ratio of the previous two values, and histograms of swapped predictions over confidence. These are tracked as the activation precision is held constant at 8 bits and the weight precision is decreased from 8 to 4 bits, allowing the quantization noise to be varied in a controlled manner. Note that the histograms are not normalized, so the area reflects the number of swapped predictions. \\~\\
It can be seen for both models that as the quantization noise increases/precision decreases, at first, the predictions to be swapped are the lower confidence ones, supporting the previously presented intuition. Moreover, even though the proportion of swapped predictions is quite high, the increase in error rate is only a small proportion of this. This is reflected in the reliability curves in Figure \ref{fig:1}, that show that low confidence predictions are also low accuracy, and supports the reasoning outlined in Section \ref{idea}. Even though post-training quantization may have caused a large number of predictions to change, the predictive accuracy of the models remains robust, as the majority of predictions that do change do not lead to an increase in error rate. As the weight precision is decreased further (and the quantization noise increases), only then do higher confidence predictions start to be swapped. The ratio of change in error rate to proportion swapped increases, and this again is reflected in Figure \ref{fig:1}, where higher confidence predictions are shown to be more accurate.
\\~\\
It is not straightforward to directly compare the two models, as the distribution of quantization noise and how it relates to the number of bits used to represent the network will be different between them. However, we can still observe in Figure \ref{fig:2} that, as more predictions in the approximate interval $[0.5, 1]$ are swapped, the ratio of change in error rate to proportion swapped increases much more quickly for ResNet50 compared to ResNet56. This can be related to Figure \ref{fig:1}, where ResNet56 is much less accurate than it is confident in this interval, whilst ResNet50 is better calibrated. This supports the idea that overconfidence improves the robustness of accuracy to quantization. 
\section{Discussion}
We have shown the novel insight that model calibration can help explain the robustness of CNN accuracy to post-training quantization. Low confidence predictions are more easily swapped post-quantization. However, if these predictions are low accuracy as well then the overall accuracy will not decrease by much. High confidence predictions will be more accurate, but more difficult to change. Moreover, we reason that overconfidence may improve robustness, as higher accuracy predictions will be more confident, and lower confidence predictions will be less accurate. We hope that this work can lay the groundwork for further analysis on the understanding of compressed neural networks. For example, further investigation should be done into how quantization precision affects noise on the logit level, as this work only examines behavior after the softmax.  
\\~\\
This result raises a potential dilemma, which is not considered in current literature. As research is increasingly moving towards producing deep learning approaches with better estimates of predictive uncertainties, better calibrated models may consequently have less robust accuracies to quantization. Interestingly, our findings, as they relate to the intrinsic calibration of a trained model, do not affect methods that improve calibration post-hoc, such as Temperature Scaling \cite{pmlr-v70-guo17a} or Deep Ensembles \cite{DBLP:journals/corr/abs-2101-05397, NIPS2017_9ef2ed4b}. However, methods applied during training, such as using a soft calibration objective \cite{karandikar2021soft} or label smoothing \cite{Mller2019WhenDL} may be affected. Thus, a natural extension of this work would be to investigate post-training quantization on models trained using these methods.

\bibliographystyle{ieeetr}
\bibliography{bibliography}


\clearpage

\appendix
\section{Appendix}\label{appendix}
\begin{figure}[h!]
    \centering
    \begin{subfigure}{0.49\linewidth}
        \centering
        \includegraphics[width=\linewidth]{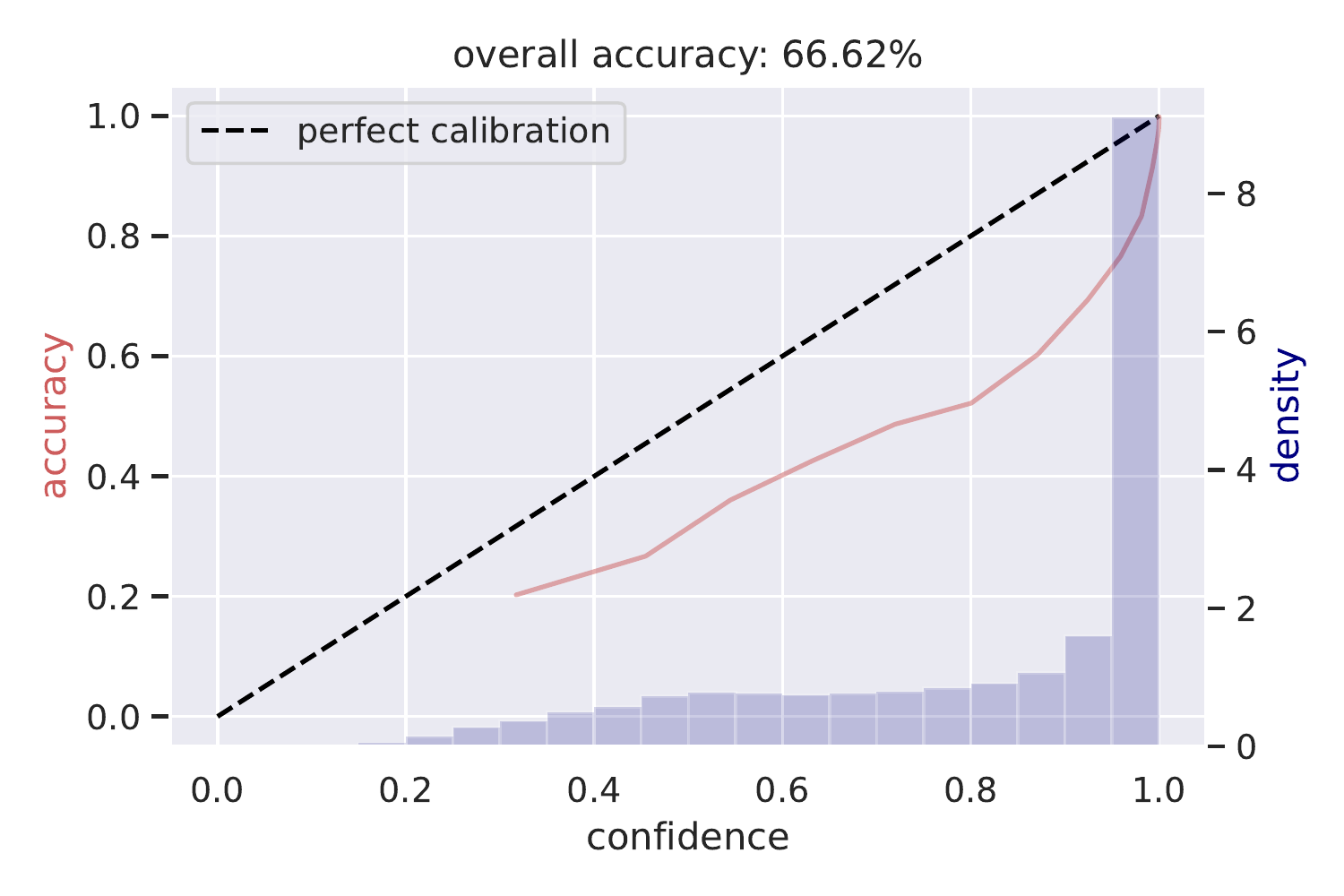}
        \caption{ResNet20, CIFAR-100}
    \end{subfigure}
    \begin{subfigure}{0.49\linewidth}
        \centering
        \includegraphics[width=\linewidth]{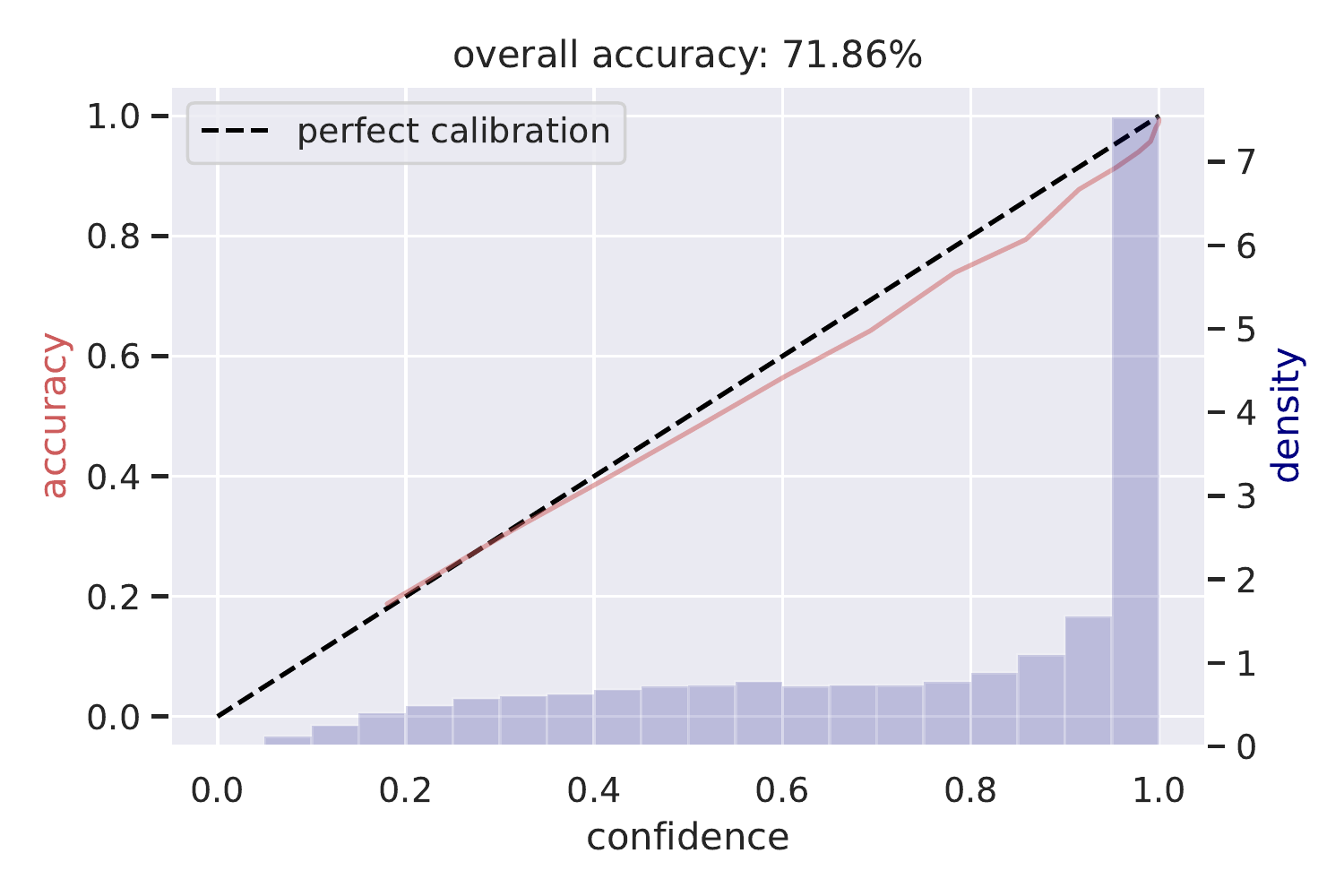}
        \caption{MobileNetV2, ImageNet}
    \end{subfigure}
    \caption{Reliability curves and confidence histograms for full precision (pre-quantization) networks on the corresponding test datasets.}
    \label{fig:3}
\end{figure}

\begin{figure}[h!]
    \centering
    \begin{subfigure}{0.49\linewidth}
        \centering
        \includegraphics[width=\linewidth]{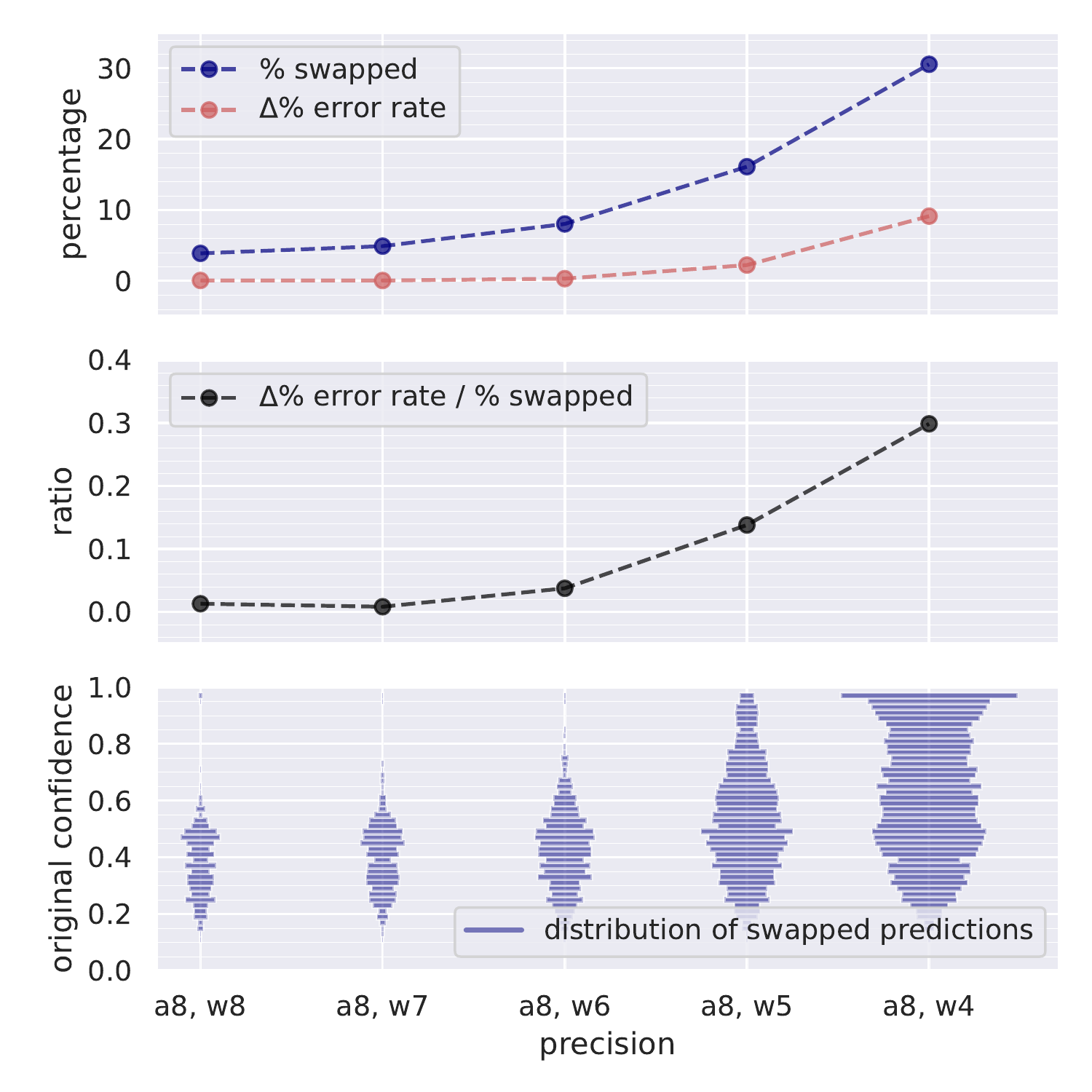}
        \caption{ResNet20, CIFAR-100}
    \end{subfigure}
    \begin{subfigure}{0.49\linewidth}
        \centering
        \includegraphics[width=\linewidth]{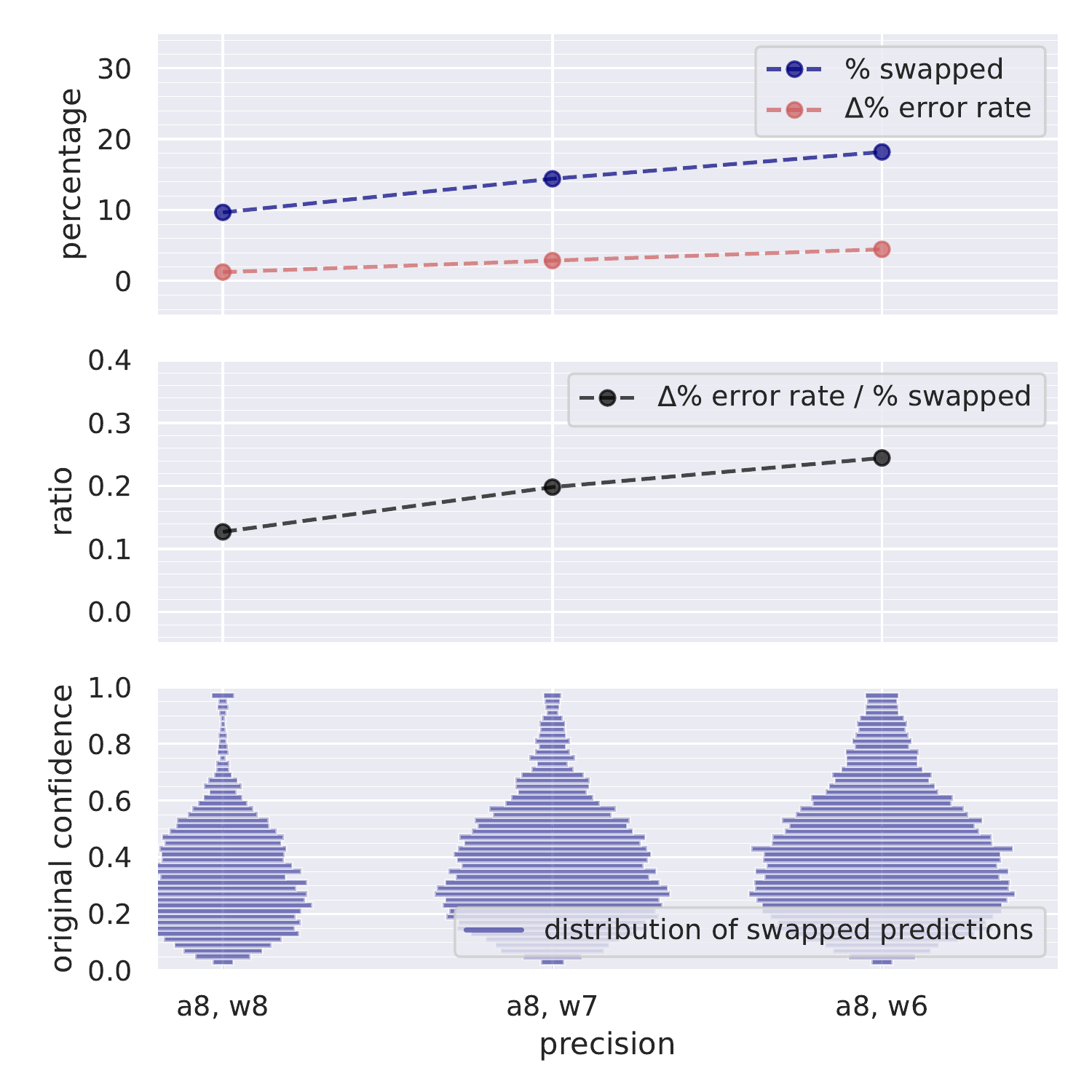}
        \caption{MobileNetV2, ImageNet, truncated for readability}
    \end{subfigure}

    \caption{Going from floating point to quantized: percentage of swapped predictions, percentage change in error rate, their ratio, and (unnormalized) histograms of swapped predictions over confidence. These are plotted as weight (w) precision is decreased whilst activation (a) precision is held constant. MobileNetV2 is truncated for readability, as the values are much higher for lower precisions.}
    \label{fig:4}
\end{figure}

Additional experimental results are provided for ResNet20 on CIFAR-100 and MobileNetV2 on ImageNet (Figures \ref{fig:3} and \ref{fig:4}). ResNet20 behaves quite similarly to ResNet56, which is to be expected as they share the same architecture. MobileNetV2 is similarly calibrated to ResNet50, but is more fragile to quantization, which is a common observation about this architecture \cite{Krishnamoorthi, nagel2021white}. However, in Figure \ref{fig:4}, its behavior is still consistent with the reasoning presented in this paper. There is just likely to be more quantization noise at the logit level for a given precision compared to ResNet50. Note that the figure is truncated for readability, as the change in values is significantly larger for lower precisions.

\end{document}